\documentclass[10pt,twocolumn,letterpaper]{article}

\usepackage{cvpr}
\usepackage{times}
\usepackage{epsfig}
\usepackage{graphicx}
\usepackage{amsmath}
\DeclareMathOperator*{\argmin}{arg\,min}
\usepackage{array}
\newcolumntype{P}[1]{>{\centering\arraybackslash}p{#1}}
\usepackage{amssymb}
\usepackage{mathtools}
\usepackage{url}
\usepackage[sorting=none, style=ieee]{biblatex}
\usepackage{mathtools}

\usepackage[numbib]{tocbibind}
\newrobustcmd{\B}{\bfseries}

\usepackage[labelfont=bf]{caption}
\usepackage{tabularx}
\usepackage{algorithm}
\usepackage[noend]{algpseudocode}
\makeatletter
\newcommand{\multiline}[1]{%
  \begin{tabularx}{\dimexpr\linewidth-\ALG@thistlm}[t]{@{}X@{}}
    #1
  \end{tabularx}
}
\makeatother
\usepackage[breaklinks=true,bookmarks=false]{hyperref}

\cvprfinalcopy 

\def\cvprPaperID{8682} 

\begin{document}

\title{A Proposed Thumbnail Selection Model}

\author{Zhifeng Yu\\
Lionsgate Entertainment Corp.\\
{\tt\small yuzhifeng1111@gmail.com}
\\
\and
Nanchun Shi\\
University of Southern California\\
{\tt\small nanchuns@marshall.usc.edu}
\\
}

\title{A Multi-modal Deep Learning Model for Video Thumbnail Selection}
\maketitle

\begin{abstract}
Thumbnail is the face of online videos. The explosive growth of videos both in number and variety underpins the importance of a good thumbnail because it saves potential viewers time to choose videos and even entice them to click on them. A good thumbnail should be a frame that best represents the content of a video while at the same time capturing viewers' attention. However, the techniques and models in the past only focus on frames within a video, and we believe such narrowed focus leave out much useful information that are part of a video. In this paper, we expand the definition of content to include title, description, and audio of a video and utilize information provided by these modalities in our selection model. Specifically, our model will first sample frames uniformly in time and return the top 1,000 frames in this subset with the highest aesthetic scores by a Double-column Convolutional Neural Network, to avoid the computational burden of processing all frames in downstream task. Then, the model incorporates frame features extracted from VGG16, text features from ELECTRA, and audio features from TRILL. These models were selected because of their results on popular datasets as well as their competitive performances. After feature extraction, the time-series features, frames and audio, will be fed into Transformer encoder layers to return a vector representing their corresponding modality. Each of the four features (frames, title, description, audios) will pass through a context gating layer before concatenation. Finally, our model will generate a vector in the latent space and select the frame that is most similar to this vector in the latent space. To the best of our knowledge, we are the first to propose a multi-modal deep learning model to select video thumbnail, which beats the result from the previous State-of-The-Art models.
\end{abstract}


\section{Introduction}

Nowadays, online videos are ubiquitous. Thumbnails, as the face of online videos, serve as the first impressions on the viewers. When a thumbnail could both well capture the content of the video and be in good quality and visually appealing, it would ultimately make the video more attractive. While manual selection for thumbnail is time-consuming and therefore impractical considering the exponential growth in the number of videos, an accurate, efficient, and automatic thumbnail selection system could ease the process for video producers, editors and platforms hosting these videos.

Previous works in selecting thumbnail have their focuses mostly on frames, while not being able to fully exploit the information carried by other modalities in the videos (e.g. metadata and audio features). Our goal is to explore the effect of incorporating all complementary information associated with the video when selecting a good thumbnail in a video. As Computer Vision and Natural Language Processing (NLP) areas are rapidly growing, more State-of-The-Art (SOTA) models have been presented and shown their advanced abilities to tackle cutting-edge problems. Bert models \cite{devlin2019bert} presented by Google were trained on large corpora of texts and made use of the novel Transformer approach to capture contextual relationship between words. The enormous size of training data of the Bert models makes them well-suited for representation learning and transfer learning. The great performance of Transformer motivates us to use this model to extract important information stored in a video's title and description. On the other hand, VGG models \cite{simonyan2015deep} are among the promising Convolutional Neural Networks (CNN) trained on ImageNet, which contains a large scale of images. The task for VGG models was to classify images. Therefore, the models are capable of generating accurate and effective representations of images. We believe they could help our model better understand the frames of the videos.

There are many video datasets available online, such as YouTude-8M. Though they provide sufficient videos for training deep networks and are abundant in variety and sources, we would like to compare our model mainly with Song \etal's proposed system \cite{song2016click}, which has been a promising tool to use on thumbnail selection task. We tried to use the same dataset for fair comparison. Unfortunately, we were only able to retrieve 700 videos provided by Song \etal's paper. As a result, we also downloaded another 471 videos from Yahoo's websites, to ensure comparable data quality. This resulted in 1,171 videos in total. The videos come with ground-truth thumbnails, which we consider as the target of our model. We then evaluated our model and Song \etal's system \cite{song2016click} on 71 test videos quantitatively.   

In summary, we propose a automatic thumbnail selection model that fully exploits the characteristics of videos using advanced deep learning techniques. Specifically, we make the following contributions in this paper:

1. We implement a frame filter model to select 1,000 frames of a video with the highest aesthetic quality.

2. We present a multi-modal deep neural networks that automatically selects the best frame that captures the content of the video by learning from frames, title, description, and audio features.

3. We collected a dataset of 471 videos including their ground-truth thumbnails and metadata, in addition to the 700 videos provided by Song \etal.

\section{Related work}

\begin{figure*} 
\centering 
\includegraphics[width=18cm, height = 7.5cm]{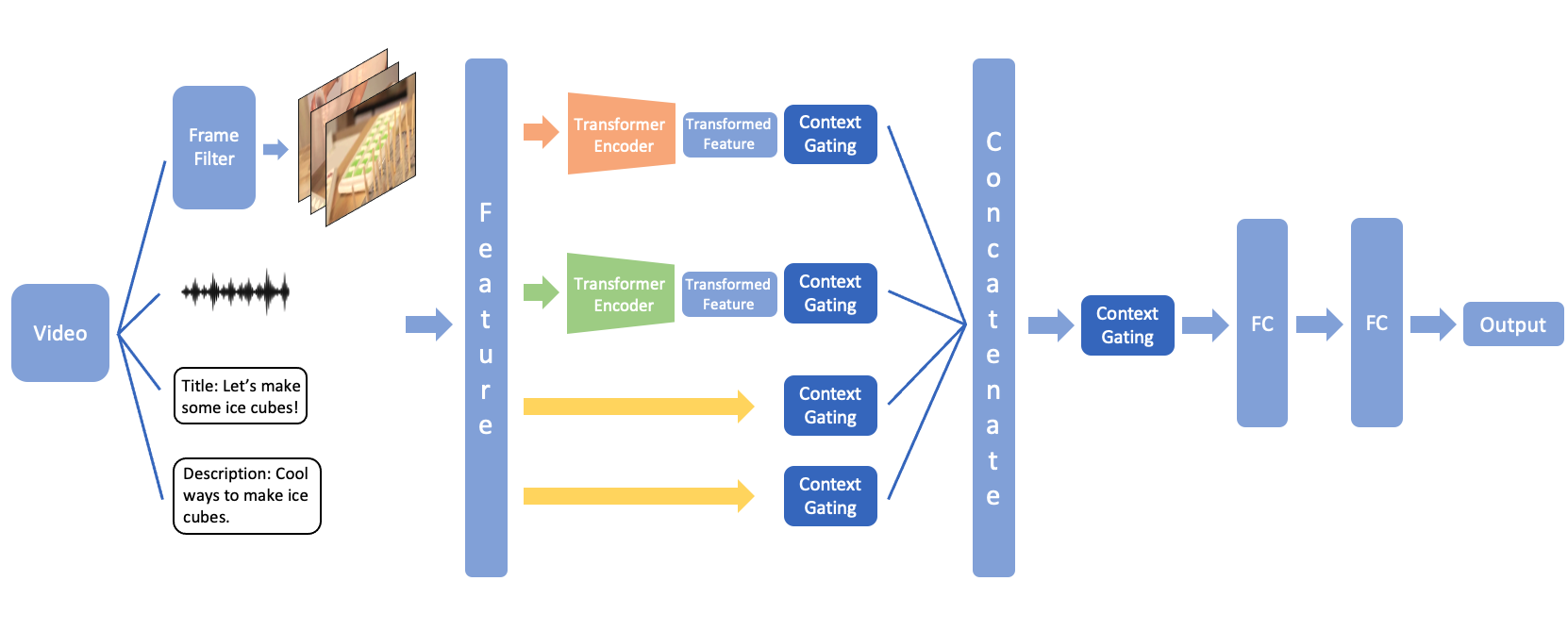} 
\caption{The architecture of our method: A video is represented by four modalities (frames, audio, title, description). The frames will go through the filter model to return a subset of frames with the highest aesthetic quality. We perform feature extractions on this subset of frames along with the other three modalities. We will then extract and apply modality-specific context information within each modality before concatenating them together. Finally, we again extract and apply global context to this concatenated vector before feeding them to the fully-connected layers and get our output}
\label{fig:ds} 
\end{figure*} 

Thumbnail selection is a relatively narrow task, so there have not been many works in this area. One of the existing systems was proposed by Song \etal \cite{song2016click}, which we consider as our benchmark in this paper. In the work, they achieve auto-selection in 3 steps. Firstly, they filter out low-quality and transitioning frames with three metrics: luminance, sharpness, and uniformity, and shot boundary detection. This approach is simple to implement and fast to compute. However, these metrics were predefined based on empirical observations and domain knowledge rather than learning from the data. This limits the model's ability to consider all potential aspects of aesthetics that can be learned from the data. In our work, we use a well-trained deep network to filter frames in the video. By doing this, we believe our model can avoid losing important frames that violate predefined thresholds and discard visually unpleasing frames that fail to be rejected. Secondly, in Song \etal \cite{song2016click}, the filtered frames are passed into subshot identification. In this step, they aim to discard near-duplicate frames in the same shot, and only keep one with the highest stillness score they define. On the contrary, we don't have explicit intervention on this. We let the model learn what's the most important to understand in the content of a video. It could be the case where several frames in the same shot contribute to the learning of motions, and therefore help our model better perceive the content. In such cases, simply removing duplicate frames to reduce processing burden might not be a good choice. Finally, with the remaining frames, they cluster them using K-means algorithm. From each cluster, they then select the frame with the highest aesthetic score (or centroids) and rank them by their corresponding cluster size. This approach is straightforward but does not effectively take the entire context of a video into account. The clusters only represent groups of closely related frames within themselves rather than being representative of the whole video. Our model learns the global context through Transformer modules, which will be discussed in detail in Section 3. By effectively utilizing both spatial and temporal information, our model can better understand the relationship between a video and individual frames. Another main difference between the system from Song \etal \cite{song2016click} and ours is that our approach utilizes multiple modalities, where we also incorporate the information from textual and audio data. Through our experiments, we found this significantly increase the accuracy of the selection. The details of these models can be found in Section 3.

There are other works in automatic thumbnail selection. Liu \etal proposed a multi-task deep visual-semantic embedding model that learns text-frame relevance in latent semantic space \cite{7298994}. Our work is similar to this in that we both take semantic (e.g., title, description, query) information into consideration and thus increases the representativeness of the selected thumbnails. However, the model takes candidate thumbnails as input, which are obtained offline. In this stage, candidates are chosen based on domain knowledge such as color entropy, motion blur and edge sharpness. Again, we believe such predefined metrics might limit the model's ability to select the diverse set of candidates and thus its practicality. Also, the semantic model computes query relevance score with individual candidate and then select final thumbnail with the highest score, whereas in our model, both within-modal context and across-modal context will be learned. This takes advantage of the temporal relationships among frames and audio and thus performs better in relevance learning. Arthurs \etal proposed an intuitive CNN classifier to label input thumbnails as "good" and "bad" \cite{Arthurs2017SelectingYV}. They make the assumptions that the number of views of a video is positively correlated to the quality of its thumbnail - the higher the number of views, the better the quality of its thumbnail has. However, the model does not take relevance with the video into account. It most likely judges based on visual quality of each candidate frame. This may work well for videos with relatively "stable" setup and backgrounds (e.g., a review video where the reviewer sitting at the same place for all the time), but cannot always work for the ones with changing scenes. In our work, we emphasize on the importance of relevance learning and achieve it by incorporating multi-modal information as well as context processing. Gao \etal pointed out that each video is associated with a central theme, which can be covered by one or more keywords \cite{5419128}. They extracted such keywords from videos' audio, tags or titles. Then candidate thumbnails will be compared to visual samples obtained from a visual database using keyword searching. This work is different from ours in that we aim to learn full context from textual data and associate candidates with each corresponding videos individually, not using an universal theme. Gu and Swaminathan developed a generalizable video reconstruction system that can be extended to various downstream tasks including thumbnail selection \cite{gu2018thumbnails}. The proposed model contains context learning using LSTM. By learning the context, the model assigns importance score to each frame that will be used for relevance selection. The authors chose not to use any semantic information as they argue in real-world scenario, there is no assurance of the quality of such semantic data. In our case, since we consider Song \etal's system \cite{song2016click} as our main benchmark, we use the extended version of the dataset Yahoo used in their paper. These videos, as compared to other sources such as YouTube, come with far more clean and relevant textual data, and we also believe in the importance of semantic information associated with a video. Therefore, we incorporated semantic information into our model.

In summary, our proposed model takes multi-modal features and learns context in both inter- and across-modal manner. To the best of our knowledge, there is no previous works using the same methodology. In this paper, we will use Song \etal's model \cite{song2016click} as our main benchmark, as it has been one of the most reliable thumbnail selection system implemented online. 


\section{Data collection}
To ensure a promising performance, we aim for a large amount of videos. The dataset we used in our experiment comes from a diverse selection of 1,171 Yahoo videos across multiple channels and under multiple categories. A category distribution is displayed in Table 1. 
\begin{table}[h!]
\centering
\begin{tabular}{l || P{0.6cm}|P{1.5cm}|P{1.5cm}} 
 \hline
  Category & \#Vid. & Mean Time (min) & Total Time (min) \\  
 \hline\hline
 Autos & 66 & 4.08 & 269.35 \\ 

 Celebrity & 46 & 2.40 & 110.20 \\

 Comedy & 16 & 3.24 & 51.78 \\

 Cute \& Inspiring & 49 & 1.73 & 84.85 \\

 Fashion \& Beauty & 59 & 2.28 & 134.27 \\

 Food & 40 & 2.75 & 109.83 \\ 
 
 Gaming & 26 & 2.90 & 75.50 \\ 

 Health \& Fitness & 43 & 2.16 & 93.02 \\ 

 International & 47 & 2.82 & 132.58 \\ 

 Makers & 243 & 2.16 & 525.73 \\ 

 Movie Trailers \& Clips & 48 & 2.23 & 106.90 \\ 

 News & 45 & 3.57 & 160.62 \\ 

 Parenting & 13 & 2.19 & 28.42 \\ 

 Sports & 66 & 1.93 & 127.07 \\ 

 TV Highlights & 49 & 3.33 & 163.08 \\ 

 Tech & 53 & 2.85 & 150.93 \\ 

 Travel & 48 & 2.83 & 135.90 \\ 
 
 Trending & 67 & 2.32 & 155.48 \\ 
 
 Unknown & 84 & 3.71 & 311.68 \\ 
 \hline\hline
 Total & 1171 & 2.64 & 3091.95 \\ 
 \hline
\end{tabular}
\\[10pt]
\caption{Category distribution of our video dataset.}
\label{table:1}
\end{table}

In order to make our model comparable with previous models, each of the videos chosen was less than 15 minutes long, as most of the models of this subject target videos of this length. The videos are 51.5 hours long in total, and the average duration is 2.6 minutes. Our videos come with high-quality thumbnails and we consider them as the ground-truth. We then split the dataset into training and test sets, consisting of 1,100 and 71 videos, respectively. The time distribution of our dataset could be found in Fig. 2. 

\begin{figure}[!ht]
\centering
\includegraphics[width=9cm, height = 5.5cm]{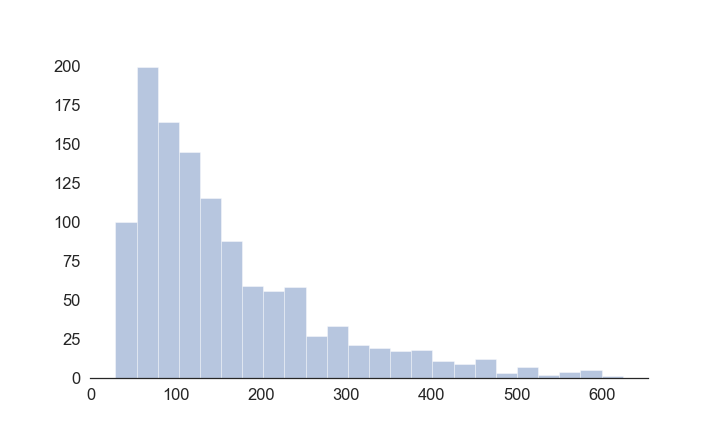} 
\caption{Video duration distribution (in seconds).}
\label{fig:ds} 
\end{figure} 

\section{Model}
When we, as human beings, try to understand a concept or an idea, we intuitively want to look for information from as many sources as practical and possible because these different information provide a holistic view on our target. We want to mimic this process in our model, and this sparks our idea of a multi-modal model in selecting a thumbnail in a video because essentially thumbnail selection can be analogized as a process of understanding a video with a representation of a thumbnail from different sources, which, in this case, are frames, title, description, and audio. In the following subsections, we will describe in details our model, which has four main modules in our model: filter model, feature extraction, context learning and fully-connected layers. A visual representation of our model is displayed in Fig. 1.
\subsection{Preprocessing}
Before the frames of a video are fed into the filter model, we have one small yet necessary preprocessing step for the frames. We have sampled 1 frame from every 9 frames in the temporal sequence of frames so as to eliminate duplicates and reduce workload. As a result of this step, the filter model will output a wider variety of frames because it will not score the almost-identical frames in close proximity in the sequence, which tend to have identical aesthetic scores. 
\subsection{Filter model}
Aesthetic quality is an important attribute of a good thumbnail since a visually pleasing image would grab attention instantly. As a result, we trained a Double-column Convolutional Neural Network (DCNN), inspired by \cite{7243357}, to score the aesthetic quality of each frame in the video and only select the top 1,000 frames. Selecting only a subset of the frames with the highest aesthetic quality also provides the benefit of lightening the computational workload in the downstream processing tasks such as learning context in the video because the model does not have to process every frame in the video. The 1,000 ordered frames are now the proxy of the population of frames in a video.
  
We found a large-scale aesthetic image dataset with 250, 000 images called AVA to help our model to learn to score images based on visual attributes \cite{6247954}. This dataset provides a distribution of scores ranging from 1 to 10 for each image. We average the scores with the weights proportional to the frequency of each score. Thus, this module is turned into a supervised learning problem with the the goal of predicting a numerical aesthetic score for each frame.

The inputs of the DCNN model are global view and local view of an image where global view is capturing the overall landscape of an image and local view is attending fine-grained details such as position, color, and brightness \cite{7243357}. The combination and balance of both provide information for aesthetic assessment that neither view alone is able to supply. Our representation of the global view, $g_w$, anisotropically resizes both height and width to the size of 224 with the shape of $g_w\in \mathbb{R}^{224\times 224\times 3}$, whereas the local view, $l_r$, is a random cropping of the image with the shape of $l_r\in \mathbb{R}^{224\times 224\times 3}$. $g_w$ and $l_r$ are passed on to and processed by two columns independently, which are convolutional blocks with convolutional and pooling layers [2]. The results of these columns, $g_w^{'}$ $\in \mathbb{R}^{64}$ and $l_r^{'}$ $\in \mathbb{R}^{64}$, are concatenated together to get the final representation $r \in \mathbb{R}^{128}$, which will be fed to a fully connected layer to get a numerical score. The architecture of our filter model is shown in Fig. 3. 

\begin{figure*} 
\centering 
\includegraphics[width=15cm, height = 7.5cm]{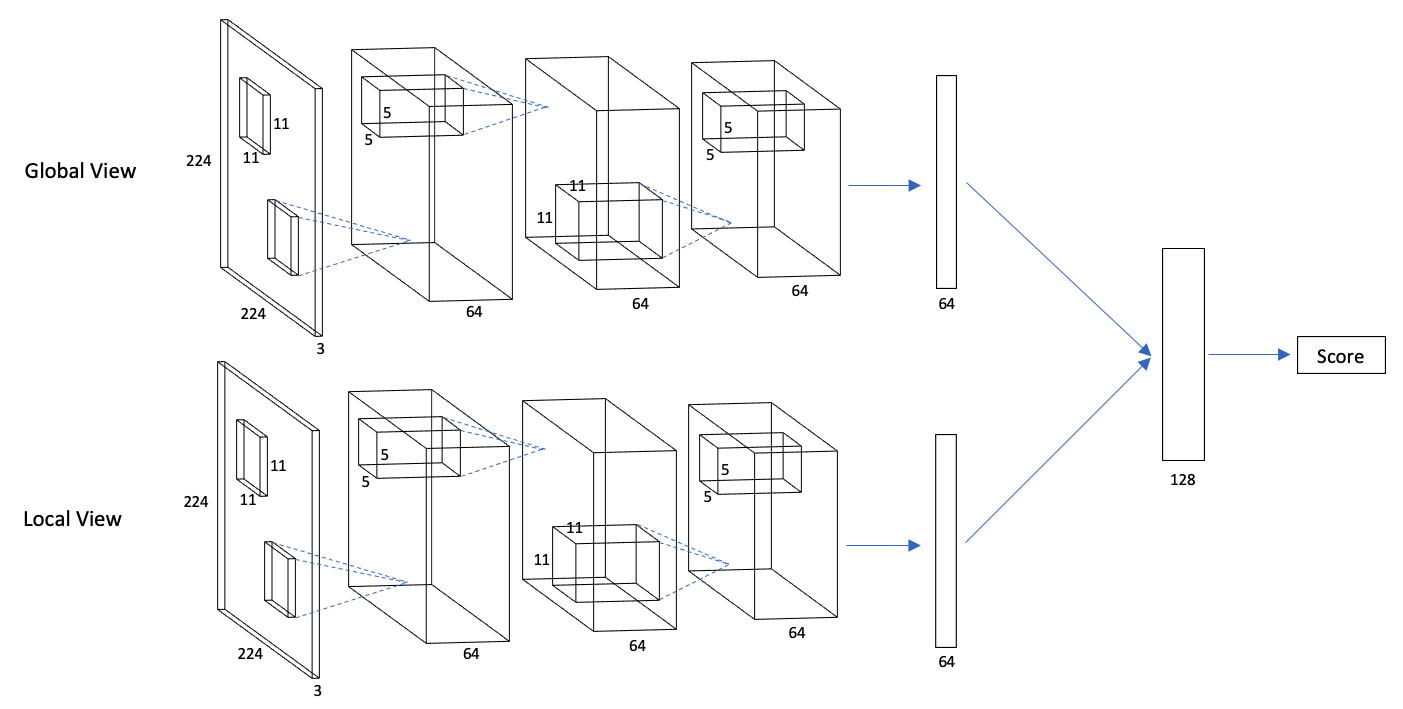} 
\caption{The structure of our filter model.}
\label{fig:ds} 
\end{figure*} 

In our experiments for DCNN, we tested with two, three, and four convolutional blocks in which each convolutional block is composed of a convolutional layer and a max pooling layer. The results of these three different models are almost identical both in speed and validation loss of Mean Squared Error (MSE). We present the results from training on 45,000 images in Table 2. Since their performances are very close, and the model with two convolutional blocks have the benefits of lower memory consumption and take up less disk space, we decided that our filter model will just utilize two convolutional blocks. We also dropped the batch normalization layers that were in \cite{7243357} because they did not improve results in our experiments. Please note that we also dropped the max pooling layer following in the second convolutional block in our final filter model. We then retrained our two-layer DCNN filter model on the entire AVA dataset. 

\begin{table}[h!]
\centering
\begin{tabular}{||P{1.5cm}|P{1.5cm}|P{1.8cm}|P{1.5cm}||} 
 \hline
  Epoch & Two-layer & Three-layer & Four-layer \\ [0.5ex] 
 \hline\hline
 1 & 0.791 & 0.737 & 0.721 \\ 
 \hline
 2 & 0.587 & 0.594 & 0.597 \\
 \hline
 3 & 0.670 & 0.661 & 0.618 \\
 \hline
 4 & 0.597 & 0.583 & 0.643 \\
 \hline
 5 & 0.598 & 0.574 & 0.574 \\
 \hline
 Time & 896s & 894s & 893s \\ 
 \hline
\end{tabular}
\\[10pt]
\caption{Validation loss of Mean Squared Error after each epoch \& time spent in total of 5 epochs (in seconds) on 45,000 images for candidate models with different convolutional layers}
\label{table:1}
\end{table}

After the filter model, we have 1,000 frames with high aesthetic quality ordered in temporal sequence for feature extraction.

\subsection{Feature extraction}
Current SOTA deep learning techniques have shown their promising power in automatically learning features. We take this advantage in our model. In the following sections, we'll introduce the feature extraction process for frames, text, and audio. 
\subsubsection{Frame embedding}
The understanding of the content of a video is crucial for thumbnail selection. A good representation learning model can group similar contents together and set apart from the very different. We expect our model to distinguish various frames so that in the later stage of comparing relevance, it could provide accurate candidates. VGG models \cite{simonyan2015deep} have been widely used in image processing tasks, and they could be easily adapted to downstream tasks. We decided to use VGG16 pre-trianed on ImageNet as our feature extractor. Specifically, we extract features from the last convolution layer with a global max pooling, resulting in 512-length vector for each frame.    
\subsubsection{Text embedding}
One of the most important aspects of auto-selection of thumbnail is its relevance to the video. Therefore, the usage of the metadata of a video, such as title and description, is desirable since they often provide rich information summarizing the content. As some studies in the past already incorporated metadata into their models, we also believe this will help our model learn the relevance between frames and videos. To achieve better representation and speed at the same time, we chose to use ELECTRA, a latest SOTA pre-trained transformer model \cite{clark2020electra}.
  
\textbf{ELECTRA} uses a new pre-training task, called replaced token detection (RTD), that trains a bidirectional model while learning from all input positions. Inspired by generative adversarial networks (GANs), ELECTRA trains the model to distinguish between “real” and “fake” input data. To achieve this, the model contains two transformers - a generator, as the precedent models like BERT, and a discriminator. The latter will take the output from the predictions of the masked tokens produced by the former as inputs, and classify them as \cite{original} and \cite{replaced}. By doing this, ELECTRA could learn from every input token, as compared to BERT-style models in which only 15\% of the tokens could be learned from. This allows the model to see fewer examples to achieve the same performance. For more information, one could refer to the ELECTRA paper \cite{clark2020electra}.

We feed title and description of each videos separately into the pre-trained model and extract features from the \cite{CLS} token, resulting in representation vectors of length 768 for each.  
\subsubsection{Audio embedding}
For the video thumbnail selection task, we found that majority of the videos involve in human interactions with the camera. In this case, the accurate representation of speech becomes pivotal for the audio embedding which will be incorporated with our final model. 

A recently developed representation network, TRILL (TRIpLet Loss network) has proven to outperform many downstream tasks for non-semantic speech representation \cite{shor2020learning}. Similar to the task of learning good representation of thumbnail embedding, TRILL was trained by minimizing closeness (L2 squared norm) in embedding space on positive/anchor pairs (closer in temporal distance) while maximizing the negative/anchor pairs. The self-supervised model was trained using proxy task of minimizing the triplet hinge loss. The hinge forces the network of not to learn a solution where it could map all outputs to zero, which results in a trivial flat loss landscape everywhere. The network was compared to many previous SOTA models and even outperformed specialized models in several tasks. 

The extraction of audio followed closely to \cite{shor2020learning}, which input takes the log mel spectrogram context window with {F} = 64 mel bands and {T} = 96 frames. Similarly, the output embedding was taken from the pre-ReLU output of the first 512-depth convolution. 

\subsection{Context learning}
After extracting features from the various modals, we have $F\in \mathbb{R}^{512\times 1000}$, $A\in \mathbb{R}^{2048\times 300}$, $t\in \mathbb{R}^{768}$, and $d\in \mathbb{R}^{768}$. We divided context learning into two parts: Within-modal context learning and Across-modal context learning. Within-modal context learning applies only to $F$ and $A$ since the very nature of these two modals is temporal. We need to distill the essential information from the temporal data into a single vector before these two modals can be used in conjunction with $t$ and $d$ to learn the final representation of the video.
\subsubsection{Within-modal context learning}
Our goal in Within-modal context learning is to aggregate $F$ and $A$ temporally to obtain a vector that contains the information across the time dimension in frames and audio. There are two considerations in our mind when selecting a model for Within-modal context learning: parallel computation and attention mechanism. 

In 3.2, we mentioned that we select the top 1,000 frames with the highest aesthetic quality as the representation of the population of frames of a video. This is not a short sequence, and the traditional sequence learning model such as Recurrent Neural Network may take a considerable amount of time to process this sequence due to its sequential nature of computation. In our model, the number of top frames is a parameter and can be increase to accommodate longer videos. As a result, we believe parallel computation in the model for Within-modal context learning is crucial.

Another key feature we want to see in the model for Within-modal context learning is flexible attention mechanism, which should not be too rigid and computationally intensive. The flexibility of this attention mechanism should confer the benefit of weighting different parts of the sequence differently based on the training dataset. Thus, this eliminates the need to pre-specify the dependencies of parts in the sequence. 

With these two requirements in mind, we found Transfomer \cite{vaswani2017attention}, a model that offers parallel computation capability in its attention mechanism process. Attention is computed as follows:   
\begin{equation}
Attention (Q, K, V) = softmax(\frac{QK^T}{\sqrt{d_k}})V\,,
\end{equation} 
where $Q, K, V$, and $\sqrt{d_k}$ are query, key, values, and scaling factor, respectively \cite{vaswani2017attention}. The essence of this process is to adjust the weights (from the matrix multiplication of $Q$ and $K^T$), which is a proxy of attention, assigned to each value based on the training set. As a result, this does not require engineers to specify the dependency of different parts in the sequence ahead of time. The parallel computation capability is made possible by the fact that the query and key vectors are packed into matrices.

In addition, this model has the extra benefits of learning attentions from different representation sub-spaces with Multi-Head Attention when multiple output values from the attention process are concatenated, rendering the model a better learner \cite{vaswani2017attention}. Multi-Head Attention is computed as follows:
\begin{equation}
MultiHead(Q, K, V) = Concat(h_1, ..., h_i)W^O,
\end{equation} 
where $h_i = Attention (QW_i^{Q}, KW_i^{K}, VW_i^{V})$, and $W_i^{Q} \in \mathbb{R}^{d_{model} \times d_k}$, $W_i^{K} \in \mathbb{R}^{d_{model} \times d_k}$, $W_i^{V} \in \mathbb{R}^{d_{model} \times d_k}$, and $W^{O} \in \mathbb{R}^{hd_v \times d_{model}}$.
Similar to \cite{vaswani2017attention}, we have h = 8 in our model. However, we have $d_k = d_h = d_{model} / h = 64$ for our frame module, $f$, and $d_k = d_h = d_{model} / h = 256$ for our audio module, $a$. In our experiments, we varied N to see its impact on performance. But we could not see noticeable improvement when we incrementally increase N from 2 to 12. Thus, we chose N = 2 in our model. Finally, in the Position-wise Feed-Forward Network, the inner-layer dimension is $d_{ff} = 128$ in our model.
Our implementation of the encoder block is adopted heavily from Tensorflow \footnote{https://www.tensorflow.org/tutorials/text/transformer}. At the end of the Within-modal context learning, we have the following representations for each module:
\begin{gather}
\nonumber frame: f \in \mathbb{R}^{512}\\
\nonumber audio: a \in \mathbb{R}^{2048}\\
\nonumber title: t \in \mathbb{R}^{768}\\
\nonumber description: d \in \mathbb{R}^{768}
\end{gather}
\subsubsection{Across-modal context learning}
Since each module is such a information-rich medium, we want to further extract context from each of the module by allowing interactions of activations and tuning the strengths of them in each module by using a layer called Context-Gating layer \cite{miech2018learnable}. The idea of this layer is to blend the information from each activation more thoroughly, and thus we will have a better representation of each module. The computation for the Context-Gating layer is as follows:
\begin{equation}
M_i^{`} = \sigma(W_iM_i + b_i) \circ M_i,
\end{equation} 
where $M_i^{`}$ is the transformed module vector, $M_i$ is the module vector, $W_i$ and $b_i$ are the trainable parameters, $i$ is the module (i.e., $f$, $a$, $t$, $d$), $\sigma$ is the element-wise sigmoid activation and $\circ$ is the element-wise multiplication \cite{miech2018learnable}. One key feature of this layer we believe is very important is the capability to recalibrate strengths of each activation in the module since $\sigma(W_iM_i + b_i) \in [0, 1]$ \cite{miech2018learnable}. This mitigates the dominant effects of some activations in the previous representation.

Finally, we concatenate the transformed module vectors, $f$, $a$, $t$, $d$, and obtain the final representation of the entire video, $V \in \mathbb{R}^{4096}$. This final representation will go through the Context-Gating layer so that different modules can interact with each other. The computation of this across-module interaction is as follows:
\begin{equation}
v = \sigma(W_vV + b_v) \circ V,
\end{equation}
where $W_v$ and $b_v$ are also the trainable parameters.
\subsection{Fully-connected layers}
After the Context Learning section, we feed $v \in \mathbb{R}^{4096}$ to two hidden fully connected layers and one output layer, each of which has 512 neurons. The output from the final layer, $o \in \mathbb{R}^{512}$, is the representation of the thumbnail generated by the model in the latent space. We compare $o$ with the 1,000 representations of frames with the top aesthetic quality in our filter model and find the one that is closest to $o$ in Mean Squared Error. That would be the thumbnail selected by our model. In mathematical terms,let's denote the set of all 1,000 frame representations as row vectors of the matrix $\bf{M}$ as: $S = \{{v_1},\ldots,{v_{1000}}\}$. Given the vector $o$, the row that returns the minimum MSE is:
\begin{equation}
v^* = \argmin_{v\in S}\|v-w\|_2^2\,,
\end{equation}
where $v^*$ is our selected thumbnail. 
\section{Experimental results}
Due to the small number of works and thus available models in this field, we only showcase the results of our model against that from the model of Song \etal \cite{song2016click} in Table 2. for quantitative evaluation. First of all, our test set of 71 videos was fed to the model provided by Song \etal \cite{song2016click} in their paper \footnote{https://github.com/yahoo/hecate} to get the top candidate for the thumbnail for each video. We call this candidate thumbnail from \cite{song2016click}, $c_{yahoo}$, and our top candidate thumbnail, $c_{ours}$. Then, we compared our results with theirs using Precision, which considers the prediction to be correct if $c_{yahoo}$ or $c_{ours}$ matches with the ground truth, $f^{*}$. However, as pointed our in \cite{song2016click}, there are many visually similar frames in the videos, we consider $c_{yahoo}$ or $c_{ours}$ as a match with $f^{*}$ if the MSE between them is below a threshold, $\theta$. As a result, True Positive of a comparison between $c_{yahoo}$ or $c_{ours}$ and $f^{*}$ is defined as:
\begin{equation}
True Positive = \begin{cases}
1, \;\;\;\;\;if \;\|f^{*}-c\|_2^2 \leq \theta\\
0, \;\;\;\;\;Otherwise

\end{cases},
\end{equation}
where $c \in \{c_{yahoo}, c_{ours}\}$.

Once we have the number of True Positives and we calculate the Precision in the test set of 71 videos of our model and \cite{song2016click} for different $\theta$. The result is presented in Table 3.

\begin{table}[h!]
\centering
\begin{tabular}{|P{1cm}||P{1.8cm}|P{1.8cm}|P{2cm}|} 
 \hline
  \multicolumn{4}{|c|}{Precision @ $\theta$} \\
  \hline
  $\theta$ & Our model & Song \etal & \% Difference\\ [0.5ex] 
 \hline\hline
 500 & \B 0.197 & 0.113 & 74.3\%\\ 
 \hline
 750 & \B 0.408 & 0.267 & 52.8\%\\
 \hline
 1000 & \B 0.648 & 0.601 & 7.8\%\\
 \hline
\end{tabular}
\\[10pt]
\caption{The comparison of results of Precision between our model and the model from Song \etal for different $\theta$, which is the Mean Squared Error threshold for us to consider the candidate thumbnail and ground truth as a match.}
\label{table:1}
\end{table}

As we can see, our model performs substantially better when $\theta$ is low, meaning that our model is better at selecting the best thumbnail when the tolerance for difference between the selected candidate thumbnail and the ground truth is low. However, as $\theta$ and thus tolerance increase, the gap in performance between our model and Song \etal \cite{song2016click} will shrink because their model is also able to select a candidate thumbnail that is similar enough to be considered a match with the ground truth.

In addition, we also want to showcase the efficacy of our model from a different perspective, from performance on a smaller training set. As mentioned before, we were only able to get 700 of 1,100 videos from the original Yahoo dataset. We trained our model on this smaller dataset and obtained results and compared with those from Song \etal \cite{song2016click}, which was trained on all 1,100 videos. The results were presented below in Table 4.

\begin{table}[h!]
\centering
\begin{tabular}{|P{1cm}||P{1.8cm}|P{1.8cm}|P{2cm}|} 
 \hline
  \multicolumn{4}{|c|}{Precision @ $\theta$} \\
  \hline
  $\theta$ & Our model & Song \etal & \% Difference\\ [0.5ex] 
 \hline\hline
 500 & 0.116 & \B 0.189 & -38.6\%\\ 
 \hline
 750 & 0.387 & \B 0.401 & -3.49\%\\
 \hline
 1000 & \B 0.689 & 0.621 & 11.0\%\\
 \hline
\end{tabular}
\\[10pt]
\caption{The comparison of results of Precision between our model and the model from Song \etal for different $\theta$, which is the Mean Squared Error threshold for us to consider the candidate thumbnail and ground truth as a match. Note that our model was trained on 700 videos whereas Song \etal was trained on 1,100 videos.}
\label{table:1}
\end{table}
Even though our model was trained on about 64\% of the original dataset, it is able to best the model from Song \etal when $\theta$ is 1,000, and the performance difference when $\theta$ is 750 is about 3.5\%. That, along with the results from Table 3, demonstrates the effectiveness of our model. 

\section{Discussion}
In our model, we are able to capture the context information from different modals and utilize these information to give us a better result. However, that comes with a trade-off to speed, especially in the filter model, because the filter has to score $\frac{1}{9}$ of all the frames before it can output the top 1,000 candidate frames with the highest visual quality. Thus, how to efficiently score frames becomes very important. We have some ideas on what can be done to reduce the amount of time used by the filter model. For instance, we can use an even simpler filter model than the current one so that it would not take as much processing time. We can also use some statistical approach to get the distribution of the positions of thumbnail and use this distribution to generate candidate positions of where the thumbnail is potentially. These ideas will shed light on directions in future research.

In addition, we believe increasing the variety of data fed into the model will improve its performance. Currently, there are four modals as inputs to the model, and they only provide high-level information of the video. Another future research area is to incorporate low-level data such as relationship of entities/objects in the video and the change in sound in the audio. However, that introduces another challenge, which is the fact that the high-level and low-level data would need to be blended together before they can be used together efficiently in the model. We hypothesize that a hierarchical model can be constructed where low-level data is first aggregated to high-level information and be combined with the existing high-level information from those four modals.

Furthermore, the reason our model is performing better than Song \etal \cite{song2016click} is its capability to extract context information. Thus, another direction of future research is to experiments different approaches of capture context. For instance, we only use a simple Transformer encoder to aggregate the frame and audio modals in the temporal dimension, which distills the context information across time. 

\section{Conclusion}
In this paper, we presented a multimodal deep video thumbnail selection system. Our system is able to select representative thumbnails for online videos while ensuring the quality of aesthetic factors. In our experiment, the system outperformed our main comparison system under varying thresholds. We also demonstrated the efficiency of our system's learning mechanism by showing it could produce better results with less training data. On the other hand, we showed the effectiveness of combining multiple modalities in video context learning. We believe such incorporation is essential for complex video understanding tasks. Moving forward, we'd like to refine our model and explore possible ways to improve our system. 
\section{Acknowledge}
We want to thank Abbass Sharif, Andrew Zhao, Luke Lu, Puhsin Huang, Jenny Zhang, and Mahesh Pandit from the program of Master of Science in Business Analytics at University of Southern California for their contributions from various points in time. Without them, this project would not be able to start and see it through the end in this extraordinary time. We appreciate their helps and efforts in the process. 






\printbibliography[heading=bibnumbered]

\end{document}